\documentclass[journal]{IEEEtran}
\usepackage{amsfonts}
\usepackage{algorithm}
\usepackage{algorithm}
\usepackage{algorithmicx}
\usepackage{algpseudocode}
\usepackage{amsmath}

\usepackage{array}
\usepackage[caption=false,font=normalsize,labelfont=sf,textfont=sf]{subfig}
\usepackage{textcomp}
\usepackage{stfloats}
\usepackage{url}
\usepackage{verbatim}
\usepackage{graphicx}
\usepackage{cite}
\usepackage{siunitx}
\usepackage{multirow}
\usepackage{booktabs}
\usepackage{threeparttable}

\usepackage{soul}
\usepackage{makecell}
\soulregister\cite7 
\soulregister\ref7  
\soulregister\label7 
\begin{document}

\title{Efficient Tactile Perception with Soft Electrical Impedance Tomography and Pre-trained Transformer}

\author{Huazhi Dong,\IEEEmembership{ Student Member, IEEE},
Ronald B. Liu,
Sihao Teng,
Delin Hu,\IEEEmembership{ Member, IEEE}, 
Sharel Peisan E,\IEEEmembership{ Member, IEEE}, 
Francesco Giorgio-Serchi,\IEEEmembership{ Member, IEEE}, 
and Yunjie Yang,\IEEEmembership{ Senior Member, IEEE}

\thanks{Manuscript received March 31, 2025; revised March 31, 2025. This work was supported in part by the European Research Council Starting Grant (Project SELECT) under Grant no.101165927, in part by the 2025 IEEE Instrumentation and Measurement Society Graduate Fellowship Award, and in part by the Worshipful Company of Scientific Instrument Makers Post Graduate Scholarship Award. (Corresponding author: Yunjie Yang)}
\thanks{Huazhi Dong, Ronald B. Liu, Sihao Teng, Delin Hu and Yunjie Yang are with the SMART Group, Institute for Imaging, Data and Communications, School of Engineering, The University of Edinburgh, EH9 3BF Edinburgh, U.K. (e-mail: huazhi.dong@ed.ac.uk; ronald.liu@ed.ac.uk;
sihao.teng@ed.ac.uk;
delin.hu@ed.ac.uk;
y.yang@ed.ac.uk). } 
\thanks{Peisan (Sharel) E is with the Institute for Bioengineering, School of Engineering, The University of Edinburgh, EH9 3DW Edinburgh, U.K. (e-mail: Sharel.E@ed.ac.uk).}
\thanks{Francesco Giorgio-Serchi is with the Institute for Integrated Micro and Nano Systems, School of Engineering, The University of Edinburgh, EH8 9YL Edinburgh, U.K. (e-mail: F.Giorgio-Serchi@ed.ac.uk).}
\thanks{Huazhi Dong and Ronald B. Liu contributed equally to this work.}
}
\markboth{ IEEE Transactions on Industrial Electronics}%
{Shell \MakeLowercase{\textit{et al.}}: A Sample Article Using IEEEtran.cls for IEEE Journals}

\maketitle

\begin{abstract}
Tactile sensing is fundamental to robotic systems, enabling interactions through physical contact in multiple tasks. Despite its importance, achieving high-resolution, large-area tactile sensing remains challenging. Electrical Impedance Tomography (EIT) has emerged as a promising approach for large-area, distributed tactile sensing with minimal electrode requirements which can lend itself to addressing complex contact problems in robotics. However, existing EIT-based tactile reconstruction methods often suffer from high computational costs or depend on extensive annotated simulation datasets, hindering its viability in real-world settings. To address this shortcoming, here we propose a Pre-trained Transformer for EIT-based Tactile Reconstruction (PTET), a learning-based framework that bridges the simulation-to-reality gap by leveraging self-supervised pretraining on simulation data and fine-tuning with limited real-world data. In simulations, PTET requires 99.44\% fewer annotated samples than equivalent state-of-the-art approaches (2,500 vs. 450,000 samples) while achieving reconstruction performance improvements of up to 43.57\% under identical data conditions. Fine-tuning with real-world data further enables PTET to overcome discrepancies between simulated and experimental datasets, achieving superior reconstruction and detail recovery in practical scenarios. PTET’s improved reconstruction accuracy, data efficiency, and robustness in real-world tasks establish it as a scalable and practical solution for tactile sensing systems in robotics, especially for object handling and adaptive grasping under varying pressure conditions. All codes and data are publicly available in Edinburgh DataShare with the identifier https://doi.org/10.7488/ds/7985.
\end{abstract}

\begin{IEEEkeywords}
Masked Autoencoder, Electrical Impedance Tomography, Tactile Sensing, Human-Machine Interaction.
\end{IEEEkeywords}

\section{Introduction}
\IEEEPARstart
{T}{actile} sensing is a foundational technology in robotics, allowing robots to interact with their environment by perceiving pressure, texture, and force through physical contact\cite{bai2023robotic,yoshimoto2018tomographic}. This capability is essential for tasks such as object manipulation, human-robot interaction, and environment exploration \cite{park2024low,yoshimoto2024design}. Despite its significance, achieving high-resolution, large-area tactile sensing remains a challenge due to inherent trade-offs among accuracy, scalability, and complexity in sensor design. Current tactile sensors typically rely on capacitive\cite{weichart2021tactile,pourjafarian2019multi}, piezoresistive\cite{lee2021flexible}, or piezoelectric\cite{lin2021skin} mechanisms. However, the spatial resolution of these systems is constrained by the number of sensor array pixels, which requires densely packed sensor arrays for higher spatial resolution \cite{wu2022tactile,hu2024two,yan2021soft}. This not only complicates fabrication and increases costs but also limits flexibility and scalability for expanded tactile sensing tasks \cite{meribout2024tactile}. 

To overcome these limitations, Electrical Impedance Tomography (EIT) has emerged as a promising alternative \cite{cui2023recent}. EIT offers several advantages over traditional methods, including the ability to achieve large-area tactile sensing with fewer boundary electrodes\cite{hardman2025multimodal,kim2024extremely,husain2021tactile}. By injecting small electrical currents through electrode pairs and measuring the resulting voltage changes, EIT reconstructs the conductivity distribution of a conductive soft layer to infer tactile inputs such as pressure or deformation \cite{dong2025learning}. However, EIT tactile reconstruction poses significant challenges in terms of accuracy and computational efficiency \cite{adler2021electrical}.
\begin{figure*}[t]
\centering
\includegraphics[width=\textwidth]{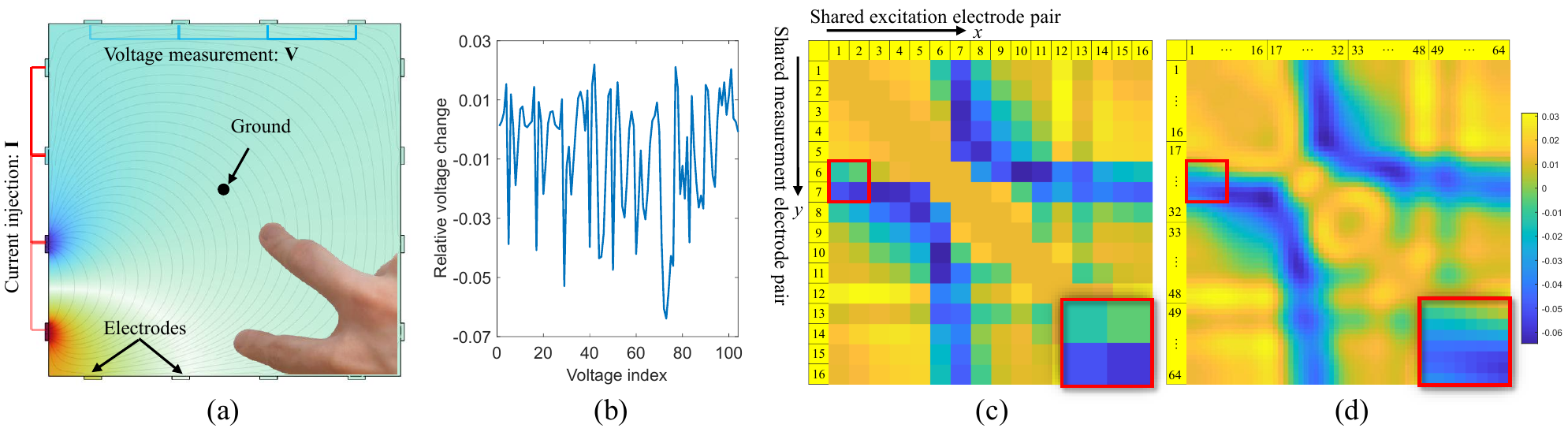}
\caption{Electrical Impedance Map (EIM). (a) Sensing principle. (b) Original measurements. (c) Construction of EIM: an example. (d) Enhanced EIM (E2IM).}
\label{fig-EIM}
\end{figure*}

Model-based methods\cite{chen2024correcting,borijindargoon2018music} have been typically employed for EIT reconstruction, solving the inverse problem non-iteratively or iteratively to estimate the conductivity distribution from voltage measurements. While these approaches do not require pre-collected datasets, they often produce suboptimal reconstruction results, especially for dynamic tactile sensing tasks. This limitation is largely attributed to EIT's inherently ill-posed and ill-conditioned inverse problem \cite{adler2021electrical}, exacerbating reconstruction quality issues. Furthermore, the accuracy of model-based approaches is inherently limited due to simplified assumptions on the governing physical model, sensor geometry and materials properties \cite{grychtol2012impact}.

Data-driven approaches, especially deep learning, have gained traction in EIT for mapping voltage measurements directly to tactile images, bypassing traditional iterative and linear methods \cite{dong2025data}. While effective, these methods rely on large annotated datasets \cite{park2021deep, park2024graph, park2022biomimetic}, which are impractical to collect at scale due to the labor-intensive nature of tactile data acquisition. As a result, training often depends on simulated data, which introduces domain gaps and reconstruction errors in real-world deployment, such as pressure inaccuracies, residual artefacts, and shape distortions \cite{ma2024pdcista}.

To mitigate this, \textit{self-supervised pretraining} has emerged as a powerful strategy, allowing models to learn generalizable features from unlabelled data and adapt to downstream tasks via fine-tuning on limited labelled samples. This paradigm has shown success in control \cite{kulkarniUnsupervisedLearningObject2019}, navigation \cite{kahnBADGRAutonomousSelfSupervised2021}, visual perception \cite{sermanetTimeContrastiveNetworksSelfSupervised2018}, and sim-to-real transfer \cite{zhaoSimRealTransferDeep2020}. Recent breakthroughs in masked prediction models, such as BERT \cite{devlinBERTPretrainingDeep2019a} and MAE \cite{he2022masked}, further highlight its scalability for high-dimensional data.

Inspired by this, we propose a self-supervised EIT pretraining framework that learns from unlabelled real-world voltages and fine-tunes on limited real voltage–tactile pairs. This enables a real-to-real learning pipeline that avoids heavy reliance on simulation and reduces domain shift errors in tactile reconstruction.

In this paper, we propose \textbf{PTET} (Pre-trained Transformer for EIT-based Tactile Reconstruction), a novel framework that tackles two core challenges in learning-based EIT tactile sensing: (1) reliance on large annotated datasets and (2) simulation-to-reality gaps. PTET employs self-supervised pretraining on unlabelled voltage data and few-shot fine-tuning on limited real-world pairs, enabling accurate tactile reconstruction with minimal annotation and reduced simulation-induced errors. This dual-phase learning strategy allows PTET to bridge the gap between experimental conditions and practical applications. Unlike traditional learning-based approaches that require extensive labelled datasets from simulation, PTET effectively operates with a small fraction of the real-world data, making it scalable and practical for real-world deployments. Our main contributions are:
\begin{itemize}
    \item We introduce the PTET model, which leverages large-scale self-supervised pretraining combined with few-shot fine-tuning to reduce dependency on large annotated datasets. This approach achieves more accurate tactile reconstructions with significantly fewer samples compared to traditional learning-based methods, making it more suitable for practical tactile systems.
    \item We develop a large-area, flexible EIT tactile sensor using a dual-conductivity layer: hydrogel for biocompatibility and a carbon black–graphite (CBG) composite for enhanced conductivity and flexibility. The sensor achieves high sensitivity and detects compressive strains as small as 1\,mm.
    \item We demonstrate the first high-resolution, learning-based tactile reconstruction trained solely on limited experimental data. This real-to-real learning pipeline eliminates simulation-related errors, advancing the practicality of EIT tactile systems.
\end{itemize}

\section{Enhanced Electrical Impedance Map} \label{sec:E2IM}
In each EIT measurement, current is injected through a pair of adjacent electrodes, and voltages are measured between neighboring pairs among the remaining electrodes, excluding the injectors (see Fig.~\ref{fig-EIM}a). 
Specifically, for a 16-electrode setup, current is injected between 16 adjacent electrode pairs, and for each excitation, voltages are measured across the remaining 13 pairs (excluding the injectors), yielding 208 measurements (16 $\times$ 13). Due to the reciprocity theorem and adjacent measurement protocol, only 104 of these are independent \cite{Yang2017}.

\begin{figure*}[t]
\centering
\includegraphics[width=\textwidth]{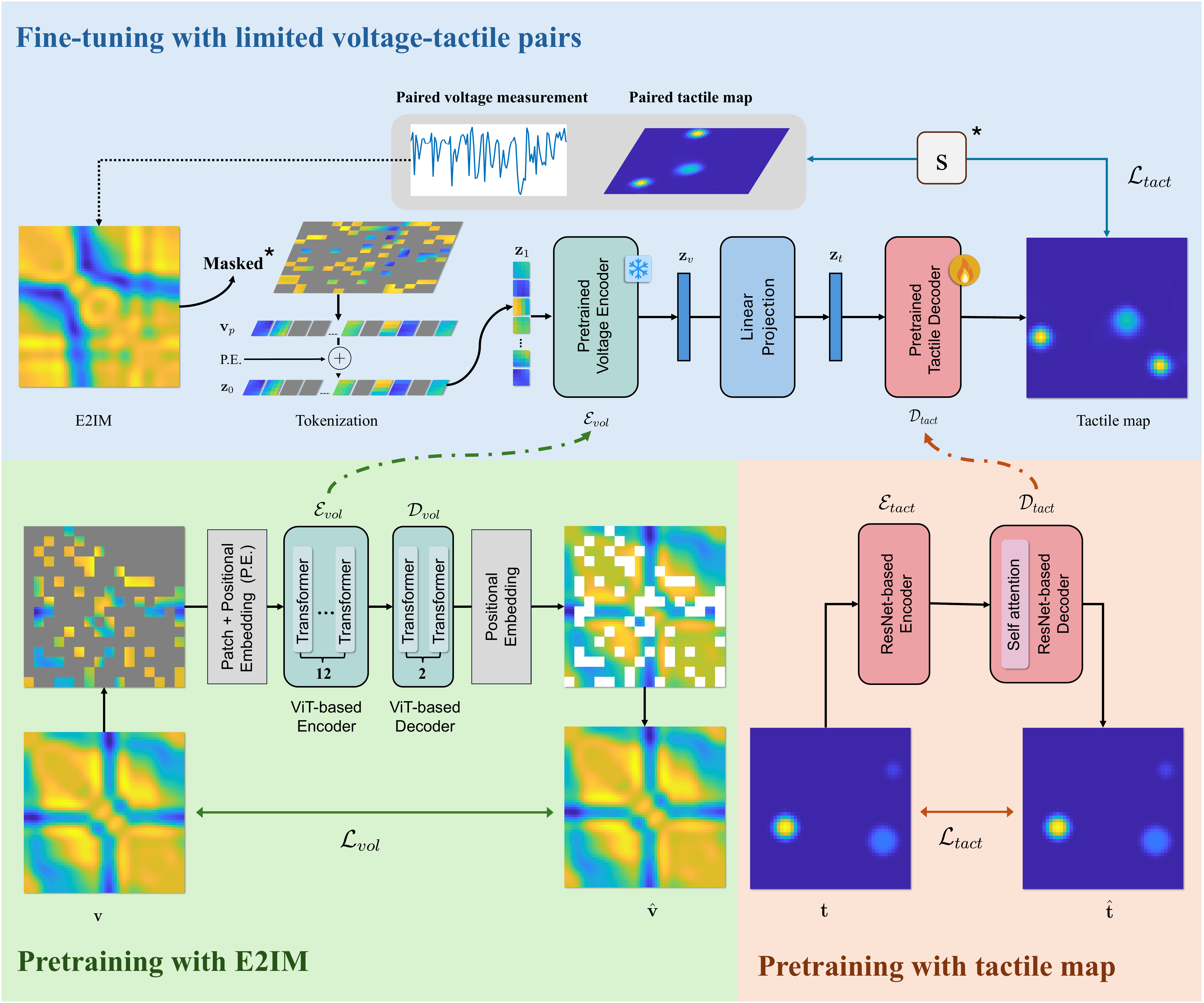}
\caption{Overview of PTET. 
Our method comprises three main processes: \textit{bottom---} 1) Pretraining with E2IM, for an effective and robust voltage encoder \(\mathcal{E}_{vol}\), through self-supervised learning by predicting masked regions in grey. 2) Pretraining with tactile map, through self-supervised learning, for a robust tactile decoder \(\mathcal{D}_{tact}\); \textit{top---} 3) Fine-tuning with limited pairs from real-world measurements, the fine-tuned model will be deployed for reconstruction (*: optional in fine-tuning). Notably, during fine-tuning,  \(\mathcal{E}_{vol}\)'s parameters are frozen, while \(\mathcal{D}_{tact}\)'s are tunable. }
\label{fig-NNArch}
\end{figure*}

To encode spatial and geometric features of EIT measurements in an image-like format, we reshape the 104\(\times\)1 voltage vector (Fig.~\ref{fig-EIM}b) into a \(16\times16\) matrix, termed the Electrical Impedance Map (EIM, Fig.~\ref{fig-EIM}c). To enhance spatial resolution, we apply bicubic interpolation, transforming the EIM into an Enhanced EIM (E2IM), where each measurement becomes a \(4\times4\) patch in a \(64\times64\) matrix (Fig.~\ref{fig-EIM}d). This augmentation improves the spatial granularity and feature richness of the original EIM, enhancing the model's capability to capture fine-grained spatial details and structural patterns, essential for achieving high-resolution tactile reconstructions and improving the robustness of the learning process. The details of generating E2IM are in the Supplementary Material.

\section{PTET: Pre-trained Transformer for EIT-based Tactile Reconstruction} \label{sec:algo}
The reconstruction of tactile representations \(\textbf{t}\) from EIT voltage measurements \(\textbf{v}\), represented in the format of E2IM, can be described by a direct mapping: 
\(f: \textbf{v} \rightarrow \textbf{t}\).

To enhance this transformation, we introduce a latent variable \(\textbf{z}\), enabling the process to be decomposed into two sequential mappings: 
\begin{equation}
    f: \textbf{v} \rightarrow \textbf{z} \rightarrow \textbf{t},
    \label{eq-vztmap}
\end{equation}
where \(p_{v}: \textbf{v} \rightarrow \textbf{z}\) maps the voltage data to the latent variable, and \(p_{t}: \textbf{z} \rightarrow \textbf{t}\) maps the latent variable to tactile representations. This decomposition compresses the information from the voltage and tactile spaces into a lower-dimensional representation, called latent space, allowing for effective feature extraction during the independent learning of \(p_{v}\) and \(p_{t}\). By leveraging self-supervised pretraining, this approach allows the models \(p_v\) and \(p_t\) to independently learn representations from the unlabeled data \(\textbf{v}\) and \(\textbf{t}\), respectively, with subsequent fine-tuning focused on aligning the latent variables. 

To implement this approach, we propose the PTET framework, built on a pretraining-based auto-encoding architecture, as illustrated in Fig.~\ref{fig-NNArch}. 
PTET comprises three key components: 
\begin{enumerate}
\item Pretraining with E2IM: In this stage, a masked autoencoder (MAE) \(\mathcal{D}_{vol}(\mathcal{E}_{vol}(\textbf{v}))\)  is employed to extract features from unlabeled voltage data \(\textbf{v}\), which is represented by E2IM. This process trains an effective and  robust encoder \(\mathcal{E}_{vol}(\textbf{v})\), functioning mapping \(p_{v}: \textbf{v} \rightarrow \textbf{z}_v\), which compresses the input \(\textbf{v}\) to a latent variable \(\textbf{z}_v\);

\item Pretraining with tactile map: In parallel with 1), an image autoencoder \(\mathcal{D}_{tact}(\mathcal{E}_{tact}(\textbf{t}))\) is utilized to learn perceptual features from unlabeled tactile data \(\textbf{t}\), represented by the 2D tactile map. This process trains a robust tactile map decoder \(\mathcal{D}_{tact}(\textbf{t})\), generating  the target output \(\textbf{t}\) from a latent variable \(\textbf{z}_t\) to implement the mapping \(p_t\);

\item Fine-tuning with limited paired voltage-tactile data: Utilizing the pretrained models from the above two steps, the voltage encoder \(\mathcal{E}_{vol}\) and tactile decoder \(\mathcal{D}_{tact}\) are aligned in a fine-tuned model, \(\mathcal{D}_{tact}(\mathcal{E}_{vol}(\textbf{v}))\), through supervised training using limited paired \(\textbf{v}\)-\(\textbf{t}\) data. This alignment focuses on the mapping between latent variables: \(\textbf{z}_v \rightarrow \textbf{z}_t\), enabling the final model to reconstruct tactile maps from the E2IM.
\end{enumerate}

PTET adopts a dual-phase training scheme: a pretraining phase with 1) and 2), and a fine-tuning phase with 3). In the pretraining phase, 1) the voltage encoder \(\mathcal{E}_{\text{vol}}\) is trained on large-scale \textit{unannotated} E2IM data, and 2) the tactile decoder \(\mathcal{D}_{\text{tact}}\) is trained separately on unannotated tactile maps. This allows both components to learn generalizable features from readily available simulation data. In the fine-tuning phase, 3) \(\mathcal{E}_{\text{vol}}\) and \(\mathcal{D}_{\text{tact}}\) are jointly refined using limited \textit{paired real-world} data to align the voltage and tactile domains. This enables accurate tactile reconstruction from real E2IM inputs. PTET employs an \textit{asymmetric} encoder–decoder architecture in all stages to improve feature representation efficiency, as detailed in the following sections.

\subsection{Pretraining with E2IM} \label{subsec:MAE}
\subsubsection{Overview}
 To pretrain E2IM \(\textbf{v}\), we designed a Transformer-based autoencoder, consisting of tactile encoder \(\mathcal{E}_{vol}\) and tactile decoder \(\mathcal{D}_{tact}\). As shown in Fig. \ref{fig-NNArch}, the voltage autoencoder trains the input E2IM \(\textbf{v}\) in a self-supervised loop \(\textbf{v} \rightarrow \hat{\textbf{v}}\). It first maps the input \(\textbf{v}\) to the latent representation \(\textbf{z}_v\): 
\begin{equation}
    \textbf{z}_v = \mathcal{E}_{vol}(\textbf{v}),
\end{equation}
then, the decoder maps the latent representation \(\textbf{z}_v\) back to the reconstructed E2IM, \(\hat{\textbf{v}}\):
\begin{equation}
    \hat{\textbf{v}} = \mathcal{D}_{vol}(\textbf{z}_v).
    \label{eq-Dvol}
\end{equation}

The pretraining loop \( \textbf{v} \rightarrow \textbf{z}_v \rightarrow \hat{\textbf{v}} \) can be framed as an optimization problem that minimizes the difference between the original E2IM \(\textbf{v}\) and the reconstructed output \(\hat{\textbf{v}}\). The loss function \(\mathcal{L}_{vol}\) is computed as the mean squared error (MSE) between the input and the reconstructed E2IM:

\begin{equation}
    \arg \min_{\mathcal{E}_{vol}, \mathcal{D}_{vol}} \mathcal{L}_{vol} = \mathbb{E}_{\textbf{v}} \left[\|\hat{\textbf{v}} - \textbf{v}\|^2 \right].
\end{equation}
This objective function aims to minimize the expectations (\(\mathbb{E}\)) of reconstruction error by optimizing the voltage encoder \(\mathcal{E}_{vol}\) and decoder \(\mathcal{D}_{vol}\).

\subsubsection{Masked autoencoder framework} 
The pretraining model for E2IM utilizes the masked autoencoder framework \cite{he2022masked} based on vision Transformer \cite{dosovitskiyImageWorth16x162021}. The key element for this model is the Transformer layer \cite{vaswaniAttentionAllYou2017}, which processes data in sequential tokens. As described in Section \ref{sec:E2IM}, voltage data \(\textbf{v}\) is represented by E2IM, a \(64 \times 64\) matrix. To process this form, E2IM data is pre-processed through \textit{masking} and \textit{tokenization}, as illustrated in \textbf{Tokenization} in Fig. \ref{fig-NNArch}. 
Specifically, for a batch of input E2IM matrices, \(\textbf{v} \in \mathbb{R}^{64 \times 64}\) is reshaped into a sequence of flattened 2D patches \(\textbf{v}_{p} \in \mathbb{R}^{N \times (P^2)}\). Here, \((P, P) = 4^2\) is the resolution of each patch, \(N =  \frac{64\times64}{4^2}  = 16 \times 16\) is the resulting number of the patches.

Then, these patches \(\textbf{v}_{p}\), preceded by an instructive class token \(\textbf{v}_{cls}\) at the start of the sequence as \(\textbf{v}_{p}^0\), are linearly projected into the Transformer layer and combined with \textit{patch and positional embeddings} \cite{ dosovitskiyImageWorth16x162021}. The resultant input voltage token \(\textbf{z}_0\), is computed as follows:
\begin{equation}
    \textbf{z}_0 = [\textbf{v}_{cls}; \textbf{v}_{p}^1\textbf{E}; \textbf{v}_{p}^2\textbf{E};\cdots;\textbf{v}_{p}^N\textbf{E}] + \textbf{E}_{pos},
    \label{eq-vol2}
\end{equation}
where \(\textbf{E} \in \mathbb{R}^{N}\) is the patch embedding vector, \(N = 16 \times 16\) is the total number of patches, and \(\textbf{E}_{pos} \in \mathbb{R}^{N+1}\) represents the positional embedding vector. 

We apply a \textbf{mask sampling} strategy to train the voltage encoder \(\mathcal{E}_{\text{vol}}\). Following \cite{he2022masked}, a high masking ratio (\(\alpha = 75\%\)) randomly masks \(4 \times 4\) \(\textbf{v}_{p}\textbf{E}\) tokens globally, maximizing information entropy and setting masked tokens as prediction targets. Only the remaining unmasked tokens \((1 - \alpha)\textbf{z}_0\) are fed into the Transformer encoder \(\mathcal{E}_{\text{vol}}\) for feature extraction:

 \begin{equation}
    \textbf{z}_1 = [\textbf{v}_{cls}; \textbf{v}_{p}^1\textbf{E}; \textbf{v}_{p}^2\textbf{E};\cdots;\textbf{v}_{p}^{(1-\alpha)N}\textbf{E}] + \textbf{E}_{pos}.
    \label{eq-vol3}
\end{equation}
 Transformer layers process the \(\textbf{z}_1\) with layers of multi-head attention \cite{vaswaniAttentionAllYou2017} (\(\mathcal{A}\)) with Layer Normalization (\(LN\)): 
 \begin{equation}
    \textbf{z}_{l}^{\prime} = \mathcal{A}(LN(\textbf{z}_{l-1})) + \textbf{z}_{l-1},
    \label{eq-vol4}
\end{equation}
where \(l=2\cdots(L+1)\) denotes the \(l_{th}\) layer, \(L= (1-\alpha)N\) represents the total number of unmasked voltage tokens \(\textbf{v}_p\). Then the outputs are processed by multi-layer perception as follows:
\begin{equation}
    \textbf{z}_{l} = MLP(LN(\textbf{z}_{l}^{\prime})) + \textbf{z}_{l}^{\prime},
    \label{eq-vol5}
\end{equation}
and the encoded latent output \(\textbf{z}_v\) from \(\mathcal{E}_{vol}\) is:
\begin{equation}
    \textbf{z}_{v} = LN(\textbf{z}_{l}).
    \label{eq-vol6}
\end{equation}
 
To complete Equation~(\ref{eq-Dvol}), the voltage decoder \(\mathcal{D}_{\text{vol}}(\mathbf{z}_v)\) uses positional embeddings \(\mathbf{E}_{pos}\) retained from tokenization to predict masked tokens and reposition them within the E2IM image (Fig.~\ref{fig-NNArch}). The final reconstruction \(\hat{\mathbf{v}}\) combines these predictions with the unmasked regions. Unlike prior methods that process voltage data \(\mathbf{v}\) sequentially \cite{park2021deep} and ignore sensor spatial layout, our 2D E2IM representation preserves electrode spatial information via \(\mathbf{E}_{pos}\), maintaining positional context throughout learning. As shown in Table \ref{tab-encpre}, increasing the dimension and size of \(\mathbf{E}_{pos}\) with E2IM significantly improved the feature extraction capability of \(\mathcal{E}_{vol}\).

\subsubsection{Implementation}
The voltage encoder \(\mathcal{E}_{\text{vol}}\) and decoder \(\mathcal{D}_{\text{vol}}\) are tailored with customized scales and hyperparameters for the \(64 \times 64\) E2IM input. The embedding dimension is set to 256 to effectively capture spatial information. An \textit{asymmetric} architecture enhances feature extraction by using 12 Transformer layers in \(\mathcal{E}_{\text{vol}}\) and a simpler decoder with 2 layers in \(\mathcal{D}_{\text{vol}}\). All layers employ 4 attention heads to ensure consistency.

\subsubsection{E2IM's Enhancement on Pretraining}
The MAE study \cite{he2022masked} showed that mask sampling strategies strongly affect model performance, though the reasons were not deeply explored. On high-semantic ImageNet images \cite{dengImageNetLargescaleHierarchical2009a}, masking less semantic structural information made reconstruction easier.

In contrast, as shown in Fig.~\ref{fig-EIM}, the EIM is a low-semantic voltage representation. Large patch masking obscures key structural features, degrading reconstruction quality, while very small patches (e.g., \(1 \times 1\)) permit easy interpolation but limit meaningful feature learning. Thus, for a \(16 \times 16\) EIM, the minimum optimal patch size \(P\) is 2.

To address this, we upsampled EIM from \(16 \times 16\) to \(64 \times 64\) with \(4 \times 4\) patches in E2IM, improving feature learning capacity. Ablation studies (Table~\ref{tab-encpre}) on 25,000 samples show E2IM reduces MSE by 99.95\% and increases SSIM by 66.07\% compared to EIM, confirming its superior representation quality.


\begin{table}[t]
\centering
\begin{threeparttable}
\caption{Encoder pretraining with EIM and E2IM}
\label{tab-encpre}
\begin{tabular}{ccc}
\hline
          & EIM       & E2IM      \\ \hline
Dimension & 16×16     & 64×64     \\
\(P\)     & 2         & 4         \\
\(\#\mathcal{E}_{vol}\) & 5.3 M  & 9.5 M \\
MSE       & 5.6466e-4 & 2.9701e-7 \\
SSIM      & 0.5991    & 0.9949    \\ \hline
\end{tabular}
\begin{tablenotes}
\footnotesize
\item \# = parameter count; M = million.
\end{tablenotes}
\end{threeparttable}
\end{table}

\subsection{ Pretraining with tactile maps} \label{subsec:tactrecons}
\subsubsection{Framework}
The tactile map \(\textbf{t}\) is set as \(48 \times 48\), i.e., \(\textbf{t} \in \mathbb{R}^{48 \times 48}\). Similar to  \(\mathcal{D}_{vol}(\mathcal{E}_{vol}(\textbf{v}))\), we designed a tactile autoencoder based on ResNet \cite{heDeepResidualLearning2016}  for self-supervised pretraining. The tactile autoencoder model consists of an encoder \(\mathcal{E}_{tact}\) and a decoder \(\mathcal{D}_{tact}\). The pretraining of tactile map \( \textbf{t} \rightarrow \textbf{z}_t \rightarrow \hat{\textbf{t}} \) can be presented as:
\begin{equation}
   \arg \min_{\mathcal{E}_{tact}, \mathcal{D}_{tact}}\mathcal{L}_{tact} = \mathbb{E}_{\textbf{t}} \left[\|\hat{\textbf{t}} - \textbf{t}\|^2 \right ],
    \label{eq-tact1}
\end{equation}
where \(\mathcal{L}_{tact}\) represents the MSE between the input and the reconstructed tactile map. The encoder \(\mathcal{E}_{tact}\) encodes \(\textbf{t}\) into a latent representation \(\textbf{z}_{t}\):
\begin{equation}
    \textbf{z}_{t} = \mathcal{E}_{tact}(\textbf{t}),
    \label{eq-tact2}
\end{equation}
The decoder \(\mathcal{D}_{tact}\) predicts from \(\textbf{z}_{t}\) to finish the self-supervised learning cycle:
\begin{equation}
    \hat{\textbf{t}} = \mathcal{D}_{tact}(\textbf{z}_{t}).
    \label{eq-tact3}
\end{equation}
\subsubsection{Implementation}
The tactile autoencoding model  \(\mathcal{D}_{tact}(\mathcal{E}_{tact}(\textbf{t}))\), following VQ-VAE's frame \cite{oordNeuralDiscreteRepresentation2018}, is stacked with convolutional ResNet blocks. Each convolutional ResNet block consists of two ResNet layers followed by a convolution layer. These blocks perform downsampling or upsampling, with the number of channels in each layer ranging from \textit{[32, 64, 128]}. 

Similar to the \textit{asymmetric} strategy in \(\mathcal{E}_{vol}\), we integrated the residual self-attention layer \cite{vaswaniAttentionAllYou2017, wangResidualAttentionNetwork2017a} only into the decoder \(\mathcal{D}_{tact}\) to enhance the tactile reconstruction capacity from the latent variable \( \textbf{z}_t\). In \(\mathcal{D}_{tact}\), a residual self-attention layer is integrated between the first two ResNet layers to enhance feature representation. The self-attention mechanism is implemented using three separate 128-dimensional linear layers to compute the Query (\(\mathbf{Q}\)), Key (\(\mathbf{K}\)), and Value (\(\mathbf{V}\)) matrices \cite{vaswaniAttentionAllYou2017}. The attention weights are calculated as follows:
\begin{equation}
    \text{Attention}(\mathbf{Q}, \mathbf{K}, \mathbf{V}) = \text{softmax}\left(\frac{\mathbf{Q}\mathbf{K}^\top}{\sqrt{d_k}}\right)\mathbf{V},
\end{equation}

where \(d_k = 128\) is the dimensionality of the Query and Key vectors. This layer refines feature extraction by selectively focusing on relevant spatial and semantic information while preserving residual connections for stable training.

\subsection{Fine-tuning with limited voltage-tactile pairs }\label{subsec:finetune}
A key challenge in learning-based tactile sensing is the limited availability of annotated real-world data, due to the high cost and manual effort required to collect and annotate precise tactile measurements under controlled conditions. To overcome this, we fine-tune a unified model composed of a pretrained voltage encoder \(\mathcal{E}_{\text{vol}}\) and tactile decoder \(\mathcal{D}_{\text{tact}}\), using a small set of voltage--tactile pairs. This enables effective adaptation to the target task despite data scarcity.

Since we have small fine-tuning (i.e., down to 2,500 paired) and large pretraining datasets (i.e., 450,000 unlabeled), it often leads to \textit{over-fitting}. To solve this problem, we propose strategies for \textit{few-shot fine-tuning for tiny datasets} as follows:

\subsubsection{Latent space compression} As outlined in Sections \ref{subsec:MAE} and \ref{subsec:tactrecons}, the deep neural network architecture of the autoencoder achieves a high compression ratio by encoding voltage and tactile data into a low-dimensional latent space. This compressed representation facilitates efficient data transformation and adaptation during the fine-tuning process. We keep 75\% random masking in the MAE encoder for lower computation and real-time gains. Masking can be optionally removed for accuracy boosts (see Suppl. Mat.).

\subsubsection{Partial tuning} Considering the significant differences in parameters between the voltage encoder \(9.54 M\) and the tactile decoder \(1.23 M\),  as shown in Fig. \ref{fig-NNArch}, we freeze all parameters in the voltage encoder \(\mathcal{E}_{vol}\) and only update the parameters in the tactile decoder \(\mathcal{D}_{tact}\) during the fine-tuning stage, as shown in Fig.\ref{fig-NNArch}. The latent variables from voltage measurement conditioning \(\textbf{z}_v\) and tactile \(\textbf{z}_t\) are connected via direct linear projection without any hidden layers:
\begin{equation}
    \textbf{z}_t = MLP(\textbf{z}_v);
    \label{eq-finetunelat}
\end{equation}
\(\mathcal{D}_{tact}\) is therefore updated by:
\begin{equation}
    \arg \min_{\mathcal{D}_{tact}} \mathcal{L}_{tact} = \mathbb{E}_{\textbf{z}_t, \textbf{t}} \left[\|\mathcal{D}_{tact}(\textbf{z}_t)-\textbf{t}\|^2 \right],
    \label{eq-Dtactloss} 
\end{equation}
where the loss function \(\mathcal{L}_{tact}\), the same as in tactile map pretraining, is calculated by MSE between the reconstructed tactile map \(\hat{\textbf{t}} = \mathcal{D}_{tact}(\textbf{z}_t)\) and the ground truth tactile map \(\textbf{t}\). 

Optionally, as illustrated in Fig.~\ref{fig-NNArch},  Eq.~(\ref{eq-Dtactloss}) can be further improved by multiplying the sensor's sensitivity matrix \(S\):

\begin{align}
    \arg \min_{\mathcal{D}_{tact}} \mathcal{L}_{tact^+} 
    &= \mathbb{E}_{\mathbf{z}_t, \mathbf{t}, S} \Big[ 
    (1-\lambda) \|\mathcal{D}_{tact}(\mathbf{z}_t) - \mathbf{t}\|^2 \nonumber \\
    &\quad + \lambda \| S \odot (\mathcal{D}_{tact}(\mathbf{z}_t) - \mathbf{t}) \|^2 
    \Big]
    \label{eq-Dtactloss_combined}
\end{align}
where \(\lambda\) denotes a control weight, \(\odot\) represents the element-wise multiplication (see Suppl. Mat.). 

\section{PTET performance on simulated test}\label{subsec:numsimu}
\subsection{Dataset}
\begin{figure}[t]
\centerline{\includegraphics[scale=0.25]{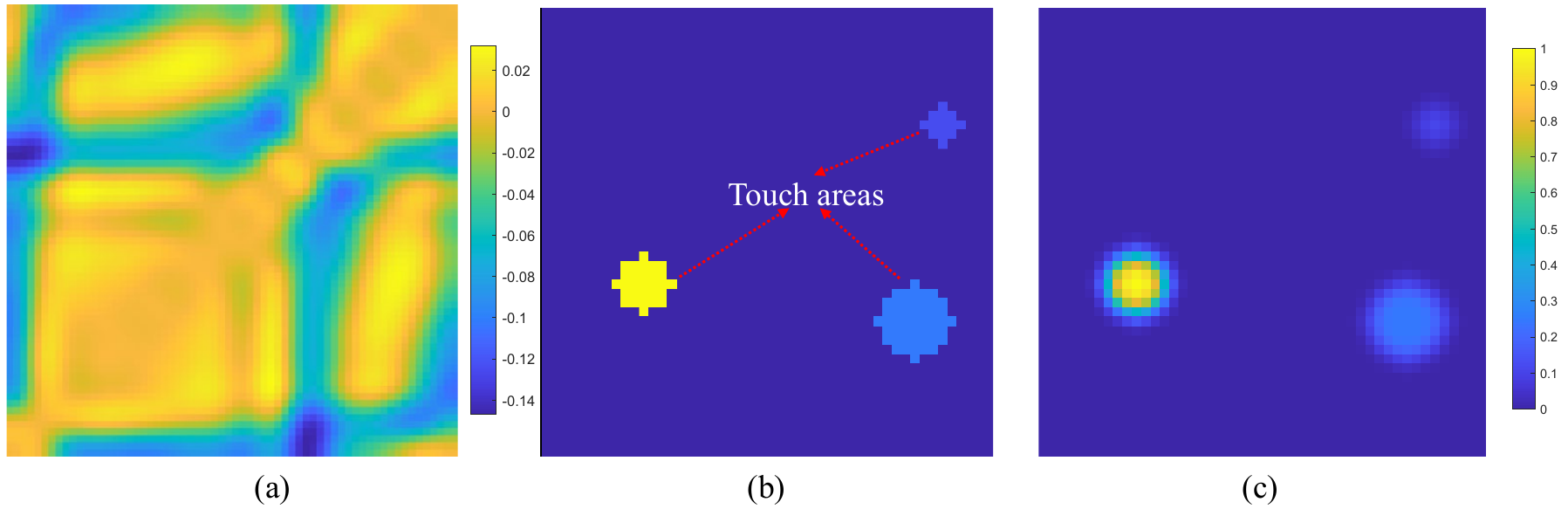}}
\caption{Example of a simulation sample. (a) E2IM (b) Ground Truth (GT) of the tactile pattern. (c) GT after Gaussian filtering. Note that we use (b) in model training and quantitative evaluation and (c) for visualization.}
\label{fig-datacollect}
\end{figure}
As shown in Fig. \ref{fig-datacollect}, the sensing region, measuring $100 \, mm\times 100 \, mm$, is divided into $48 \times 48$ pixels with a background conductivity of 0.00312 S/m. Circular areas with diameters ranging from 7.5 mm to 27.5 mm are used to represent tactile interactions, emphasizing both position and intensity. The conductivity within these touch regions is randomly varied between 0.05 and 2 times the background value to simulate diverse touch effects. The dataset generated comprises 500,000 samples, with each of the five subsets containing 100,000 samples. Each sample includes one E2IM and corresponding Ground Truth (GT). These subsets correspond to different numbers of touch regions, ranging from 1 to 5 touch areas per sample. Each subset is further partitioned into training, validation, and testing datasets using an 18:1:1 ratio.  This results in a total of 450,000 samples for training, 25,000 for validation, and 25,000 for testing. All data were generated using COMSOL Multiphysics and Matlab. The detail of the framework of the 2D EIT simulation dataset generation is shown in Supplementary Fig. 1. As shown in Supplementary Fig. 2, the waveform shapes of the reference voltage signals from simulation and real-world measurements are highly consistent.

\subsection{Comparison Algorithms}\label{subsec:compalgo}
To evaluate the effectiveness of PTET, we compared it with the state-of-the-art (SOTA) DNN-based algorithm \cite{park2022biomimetic}, another autoencoder-based model designed for EIT-based tactile sensing. Additionally, we developed a supervised variant of PTET, termed PTET-SL, which employs the same neural network architecture as the fine-tuned model but is trained directly on paired datasets without pretraining. This modification aims to demonstrate the reliance of supervised models on large datasets while showcasing the superior performance of PTET. For a fair comparison, all algorithms utilized a unified mean squared error (MSE) loss function. The parameters for the SOTA algorithm were the same as in \cite{park2022biomimetic}, with only minor adjustments made to accommodate the input dimensions of our dataset. Both PTET and PTET-SL were trained using identical parameters (see Table \ref{tab-trainparams}).
\begin{table}[t]
    \centering
    \caption{Training parameters}
    \setlength{\tabcolsep}{3pt} 
    \renewcommand{\arraystretch}{1.2} 
    \resizebox{\columnwidth}{!}{ 
    \begin{tabular}{lccc} 
    \toprule
    Parameter & Pretraining for \(\mathcal{E}_{vol}\) & Pretraining for \(\mathcal{D}_{tact}\) & Fine-tuning \\ 
    \midrule
    Optimizer type & Adam with weight decay & Adam with weight decay & Adam \\
    Base learning rate & 1.5e-4 & 1.5e-4 & 1e-4 \\
    Weight decay & 0.05 & 0.05 & N/A \\
    Learning rate scheduler type & Cosine annealing & Metric monitoring & Cosine annealing \\
    Learning rate reducing factor & 0.5 & 0.5 & 0.5 \\
    Learning rate reducing patience & N/A & 10 & 10 \\
    First momentum estimate (\(\beta_1\)) & 0.9 & 0.9 & N/A \\
    Second momentum estimate (\(\beta_2\)) & 0.95 & 0.95 & N/A \\
    Warmup epoch & 200 & 200 & 10 \\
    Total epoch & 2000 & 2000 & 2000 \\
    Early stopping patience & 200 & 200 & 200 \\
    Dropout rate & N/A & N/A & 0.5 \\
    Sensitivity loss weight (\(\lambda_s\)) & N/A & N/A & 0.3 \\
    \bottomrule
    \end{tabular}
    }
    \label{tab-trainparams}
\end{table}

\begin{figure}[t]
\centerline{\includegraphics[scale=0.35]{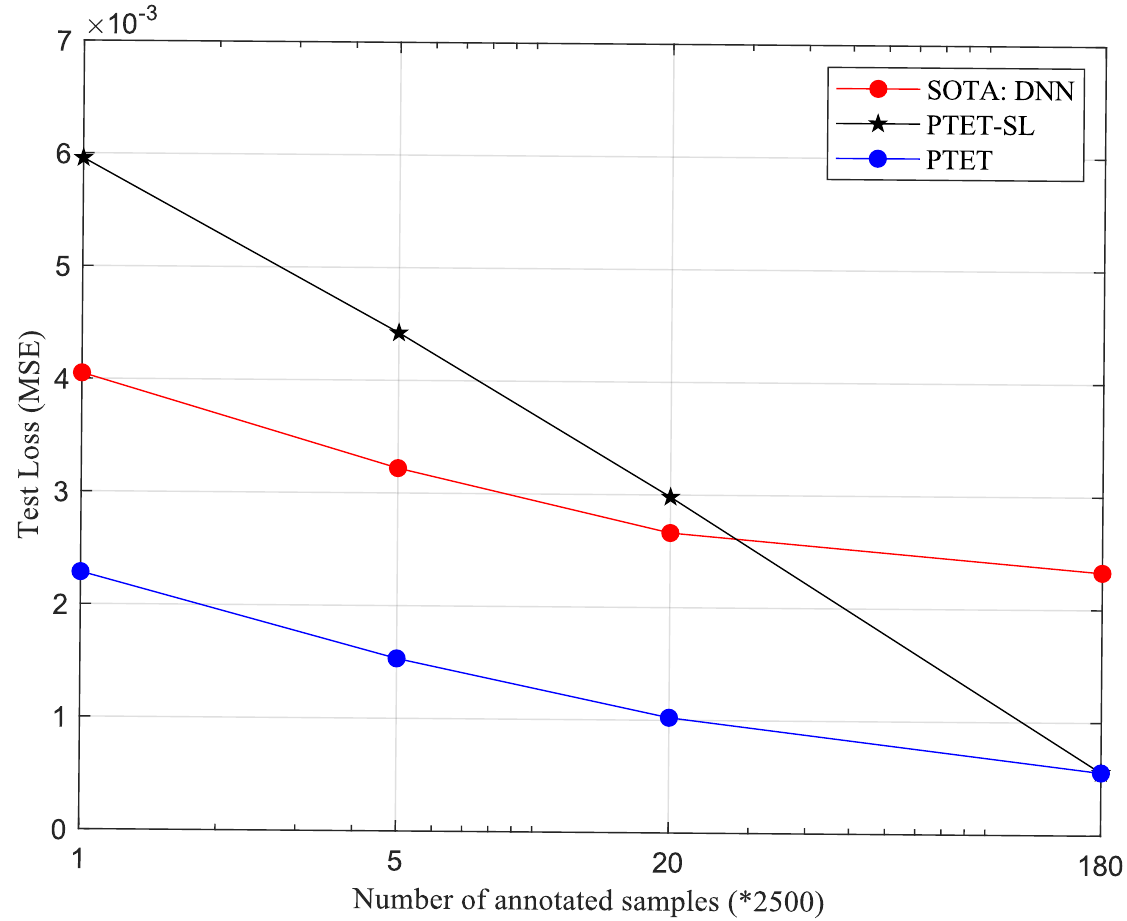}}
\caption{Test loss with different numbers of annotated samples.}
\label{fig-ComparisonAlgorithms}
\end{figure}

We divided the training dataset into 180 groups, each containing 2,500 samples. The three algorithms (PTET, PTET-SL, and SOTA DNN) were trained using subsets of the data consisting of 1, 5, 20, and 180 groups, respectively. The trained models were evaluated on a test dataset comprising 25,000 samples (see test loss in Fig. \ref{fig-ComparisonAlgorithms}). The PTET model achieved a test loss of 0.002285 using only 2,500 annotated samples. Remarkably, this matches the performance of the SOTA algorithm, which required 450,000 annotated samples (180 times of PTET) to achieve a test loss of 0.002329. Furthermore, PTET significantly outperformed PTET-SL when using less annotated samples (e.g., below 50,000 samples).  

\subsection{Quantitative evaluation}
We evaluated the tactile reconstruction performance using three quantitative metrics: Relative Error (RE), Peak Signal-to-Noise Ratio (PSNR) and Correlation Coefficient (CC). These results, summarized in Table \ref{tab-QuantitativeMetrics}, provide a comprehensive assessment of tactile reconstruction quality from different perspectives. All metrics were calculated as the mean values over the entire testing dataset (25,000 samples), ensuring a robust and reliable evaluation of model performance. Across all metrics, PTET demonstrated superior performance. Remarkably, PTET achieved superior performance using only 0.56\% of the annotated samples required by the SOTA DNN model (2,500 vs. 450,000) or just 5\% (2,500 vs. 50,000) of the annotated samples required by the PTET-SL, while consistently delivering better results.
This highlights PTET's efficiency and effectiveness in scenarios with limited annotated data, a significant advantage in real-world applications where labelled datasets are scarce or costly to obtain.

\begin{figure*}[t]
\centerline{\includegraphics[scale=0.5]{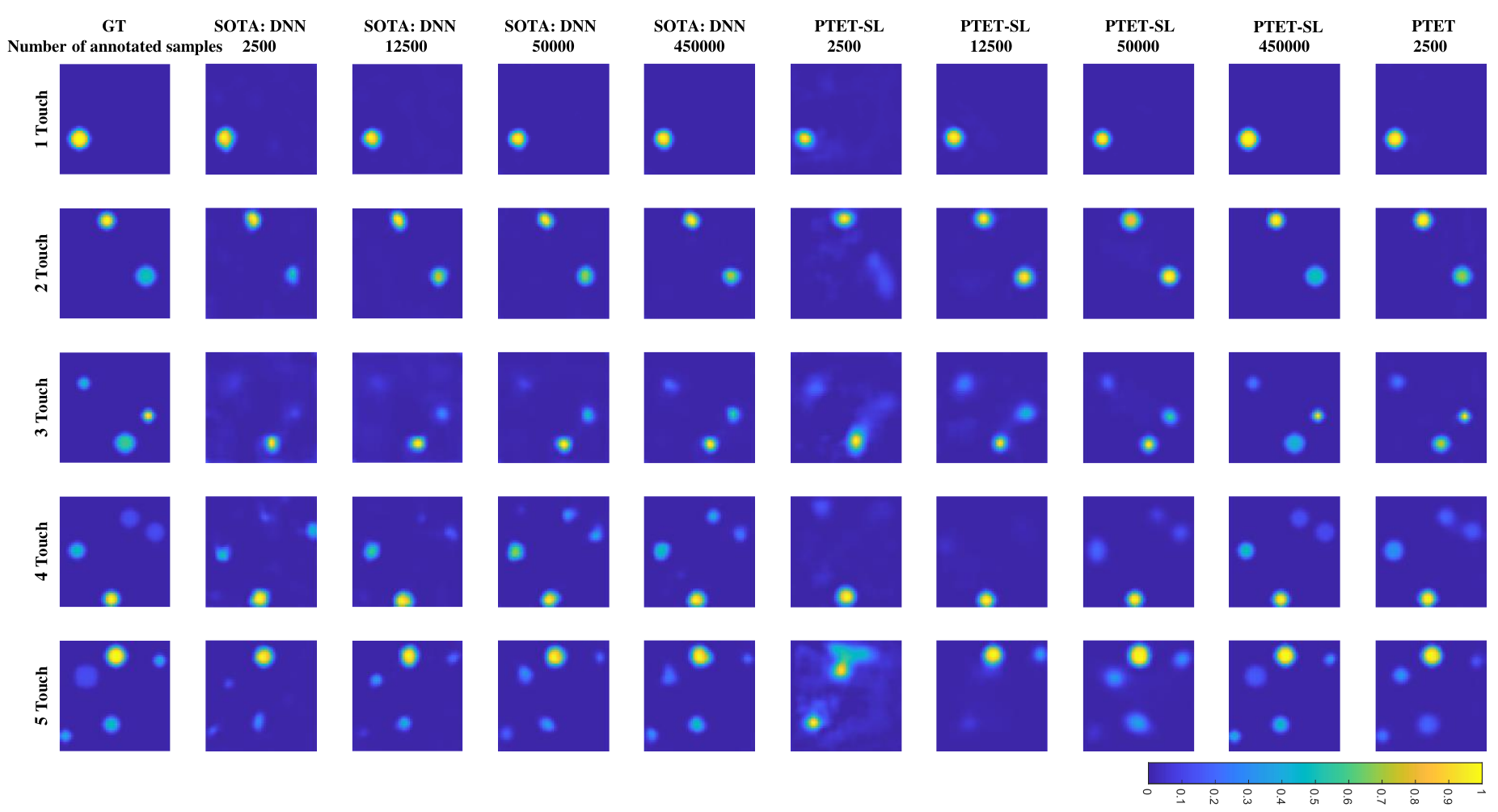}}
\caption{Representative tactile reconstruction results. For better visualization, all results were refined (Gaussian filtering) and normalized. }
\label{fig-SimulationResults}
\end{figure*}

To further demonstrate the superiority of PTET, we visually compared the reconstruction results for five representative phantoms from the 25,000 test samples (Fig. \ref{fig-SimulationResults}). The results highlight PTET's ability to accurately reconstruct positions, shapes and magnitudes of touches across varying touch scenarios, including complex cases with multiple touch points (1–5 touch points).

In contrast, the SOTA DNN, even with 450,000 annotated samples, struggled to capture fine details in scenarios with higher touch complexity, resulting in inaccuracies and noise artefacts. Similarly, PTET-SL trained with 50,000 annotated samples failed to match PTET's precision and noise reduction capabilities. The self-supervised PTET model, trained with only 2,500 annotated samples, consistently outperformed both alternatives, delivering sharper, more accurate reconstructions and further validating the effectiveness of the self-supervised learning approach.
\begin{table}[t]
\centering
\caption{Quantitative Metrics}
\label{tab-QuantitativeMetrics}
\setlength{\tabcolsep}{4pt} 
\renewcommand{\arraystretch}{1.2} 
\resizebox{\columnwidth}{!}{ 
\begin{tabular}{lcccc} 
\toprule
\textbf{Quantitative Metrics} & \textbf{RE} & \textbf{PSNR} & \textbf{CC} & \textbf{MSE ($\times 10^{-2}$)} \\ 
\midrule
SOTA (2,500)   & 0.6479 & 31.1579 & 0.6203 & 0.4049 \\ 
SOTA (12,500)  & 0.5811 & 32.2255 & 0.6950 & 0.3220 \\ 
SOTA (50,000)  & 0.5213 & 33.1799 & 0.7490 & 0.2663 \\ 
SOTA (450,000) & 0.4941 & 33.6880 & 0.7773 & 0.2329 \\ 
PTET-SL (2,500)   & 0.7533 & 29.6557 & 0.5128 & 0.5957 \\ 
PTET-SL (12,500)  & 0.6440 & 31.2573 & 0.6597 & 0.4422 \\ 
PTET-SL (50,000)  & 0.5309 & 33.8749 & 0.7627 & 0.2981 \\ 
PTET (2,500)      & \textbf{0.4936} & \textbf{34.3308} & \textbf{0.7985} & \textbf{0.2285} \\ 
\bottomrule
\end{tabular}
}
\vspace{3mm}
\raggedright
\footnotesize{Note: The best results are highlighted in bold.}
\end{table}

\section{Sensor Fabrication and Characterization}
This section describes the EIT-based tactile sensor design, fabrication and characterization, which is utilized to further validate the real-world performance of PTET.
\subsection{Sensor Fabrication}
\begin{figure*}[t]
\centerline{\includegraphics[scale=0.45]
{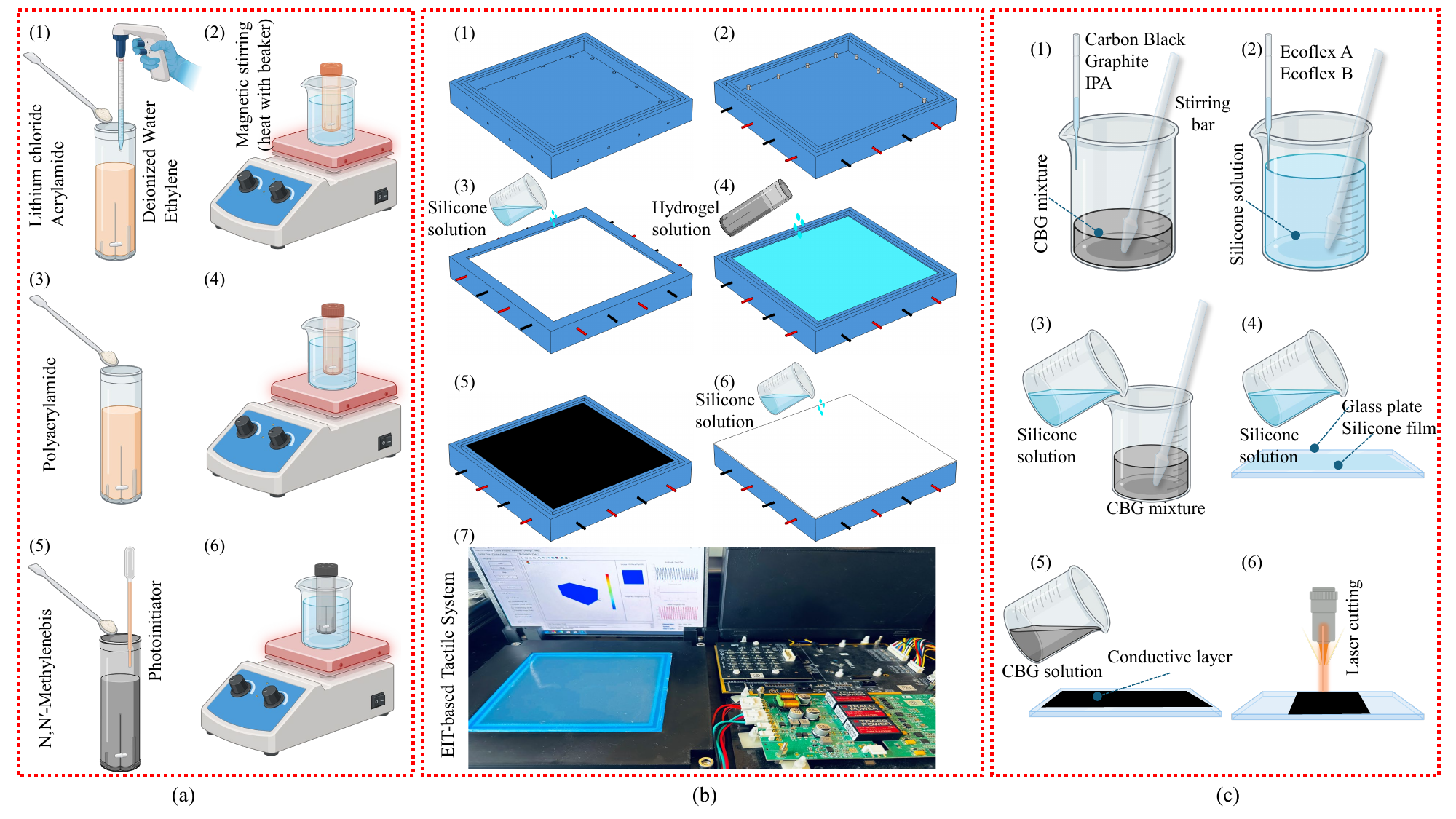}}
\caption{Sensor fabrication process. (a) Hydrogel fabrication process. (b) Steps of sensor fabrication. (c) Fabrication process of the conductive layer.}
\label{fig-SensorFabrication}
\end{figure*}
We designed a dual-conductivity layer tactile sensor combining a hydrogel layer and a CBG conductive layer, leveraging the complementary properties of each material. Hydrogels are widely utilized in electronic skin applications due to their desirable electrical properties, flexibility and biocompatibility \cite{park2022biomimetic}. The CBG layer with lower resistance improves EIT-based sensor performance. Fig. \ref{fig-SensorFabrication} shows the fabrication process and design of the EIT-based tactile sensor. The sensor dimensions are $120 \, mm\times 120 \, mm\times 10 \, mm$. A step-by-step description of the fabrication process is available in the Supplementary Material.

The dual-layer design combines the flexibility and biocompatibility of the hydrogel with the enhanced conductivity and structural integrity of the CBG layer, maximizing signal response under applied pressure and making it well-suited for applications in electronic skin and human-machine interfaces. To enhance durability and maintain long-term functionality, the entire hydrogel layer was encapsulated with a stretchable silicone elastomer to serve as a physical barrier against moisture loss \cite{zhu2020}.

\subsection{Sensor Characterization}
\begin{figure}
\centerline{\includegraphics[scale=0.4]{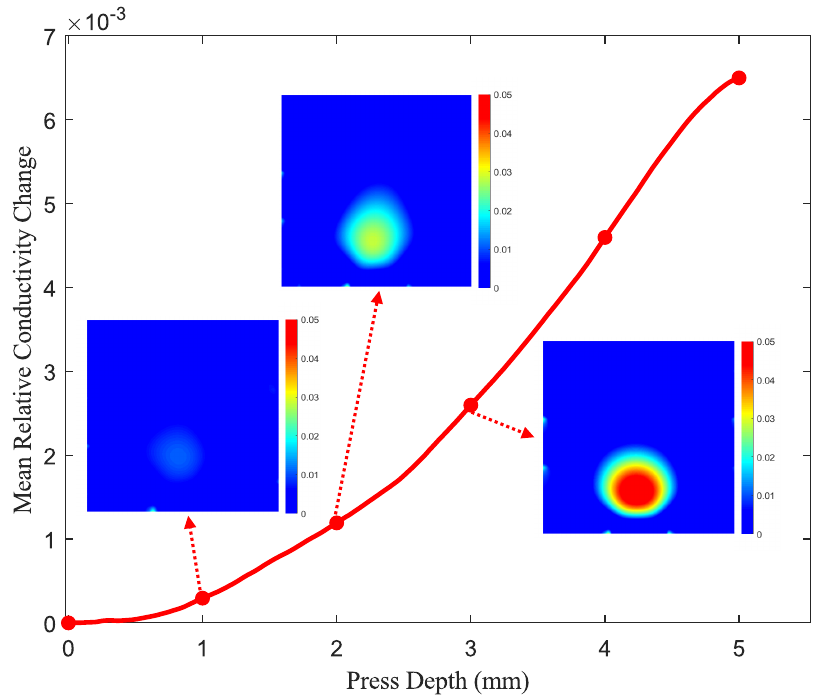}}
\caption{Sensor sensitivity analysis. The conductivity distribution is generated using Tikhnov Regularization \cite{lionheart2004eit}.}
\label{fig-SensorCharacterization1}
\end{figure}

To evaluate the sensitivity of the developed sensor, we conducted an experiment where the sensor was subject to a continuously increasing localized pressure with strains ranging from 0 mm to 5 mm, as illustrated in Fig. \ref{fig-SensorCharacterization1}. The results demonstrate a clear increase in the conductivity distribution as the compressive strain increases. Conductivity distribution images were generated for compressive strains of 1 mm, 2 mm, and 3 mm. At a compressive strain of 1 mm, the changes in conductivity are subtle yet detectable, while at 2 mm, the effect becomes more pronounced. These results demonstrate the sensor's sensitivity and capability for accurate and responsive tactile detection.

To evaluate the sensor’s durability and repeatability, we performed a cycling test involving 300 pressing cycles with compressive strains from 2 mm to 5 mm. The results show that the sensor maintains stable output throughout the cycles, with consistent trends in signal changes (see Supplementary Fig. 3). During this test, a dynamic baseline approach was adopted to mitigate slight signal drifts attributed to the lagging effect of the sensor after prolonged pressing. These results confirm the sensor's robustness and suitability for applications requiring long-term, repeated tactile interactions.

All experiments in this study were conducted using the three-axis linear robot (IGUS, DLE-RG-0001), which ensured precise control over the position and depth of the presses (see Supplementary Fig. 4). All experiments were carried out under ambient laboratory conditions, without additional environmental control, and the pressing was performed at a constant strain rate of approximately 1 mm/s. An in-house developed EIT system was utilized for data acquisition as described in \cite{Yang2017}.

\section{PTET performance in physical experiments}
\subsection{Real-world Dataset}
A diverse real-world dataset was created using 84 custom-fabricated touch tips produced by a 3D printer. The design of the touch points, including their positions and sizes, was randomly generated, with pressing depths precisely controlled between 2 mm and 5 mm. 

We collected a total of 5700 annotated samples, distributed as follows: 613 samples with 1 touch point, 1,063 with 2 touch points, 1,043 with 3 touch points, 1,004 with 4 touch points, 995 with 5 touch points, 493 with annular touch and 489 with L-shaped touch. Each sample consists of a 104-dimensional vector of voltage measurements, which is transformed into a \(64\times64\) E2IM matrix and a \(48\times48\) image representing the press map. The samples were randomly shuffled and split into three datasets in a 7:2:1 ratio for training, validation, and testing. In addition, we augmented the training set with 30 frames of pure noise (i.e., no-touch) data to better simulate sensor noise and improve model robustness. To expedite training, we reused the pre-trained models from the simulation phase for both the PTET algorithm and the baseline DNN algorithm. This approach is justified because the pretraining model only captures the fundamental features of voltage measurements and pressure maps.

\subsection{Quantification of Reconstruction}
\begin{figure*}[t]
\centerline{\includegraphics[scale=0.5]
{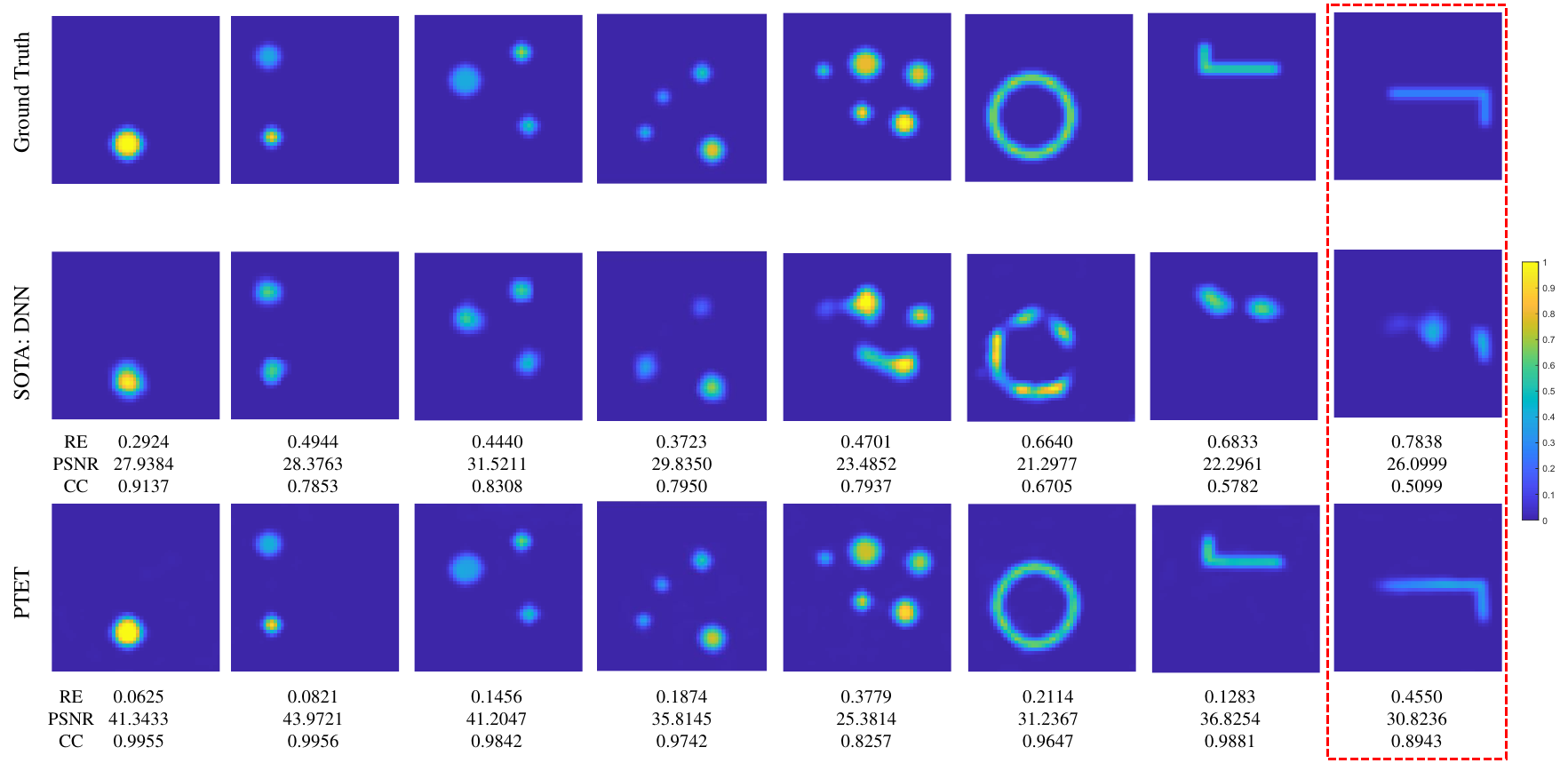}}
\caption{Quantification of Reconstruction. The red dash-line shows generalization to an unseen L-shaped sample. Results were refined using Gaussian smoothing.}
\label{fig-experimentrecon1}
\end{figure*}
We conducted a quantitative evaluation of the test dataset consisting of 570 samples, with results summarized in Table \ref{tab-QuantitativeMetricsForTestdata}. The PTET model demonstrated far superior tactile reconstruction performance across all metrics compared to SOTA DNN. Specifically, PTET reduced the RE by 21.92\%, improved the PSNR by 11.15\%, increased the CC by 7.96\%, and remarkably reduced MSE by 63.88\%.  To visually compare model performance, we selected seven representative frames from the test Dataset, as displayed in Fig. \ref{fig-experimentrecon1}. The DNN model performs comparably to PTET on simpler Phantoms (i.e., Phantoms 1-3), but fails to achieve similar tactile reconstruction quality on more complex Phantoms (Phantoms 4-7).  Additionally, we tested model generalization using a set of unseen L-shaped data with sizes not included in the training set. The reconstruction results for this sample (highlighted by red dash-line) achieved quantitative metrics comparable to the test dataset average, demonstrating the strong generalizability of PTET.

Notably, the DNN model chosen for comparison performed exceptionally well in \cite{park2022biomimetic}, as it was trained on 583,360 annotated simulation samples. However, in our experiments, with only 3,990 annotated data (around 1/150) for training, the DNN model's performance was significantly inferior. This suggests the superior performance of PTET when trained on a limited amount of annotated data, demonstrating its robustness and efficiency in data-scarce scenarios.
\begin{table}[t]
\centering
\caption{Average Quantitative Metrics on Test Set}
\label{tab-QuantitativeMetricsForTestdata}
\setlength{\tabcolsep}{6pt} 
\renewcommand{\arraystretch}{1.2} 
\begin{tabular}{lcccc} 
\toprule
\textbf{Quantitative Metrics} & \textbf{RE} & \textbf{PSNR} & \textbf{CC} & \textbf{MSE ($\times 10^{-2}$)} \\ 
\midrule
PTET          & \textbf{0.3388} & \textbf{32.5171} & \textbf{0.8772} & \textbf{0.1914} \\ 
SOTA: DNN     & 0.4339          & 29.2558          & 0.8125          & 0.5299          \\ 
\bottomrule
\end{tabular}
\vspace{1em} 
\end{table}

\subsection{Performance on Complex Touches}
To evaluate the model's effectiveness in tactile reconstruction during practical operations, we tested it with a series of complex touch scenarios. These included configurations involving multiple fingers, annular touch, annular touch with finger, and L-shaped touches. 

\begin{figure}
\centerline{\includegraphics[scale=0.4]{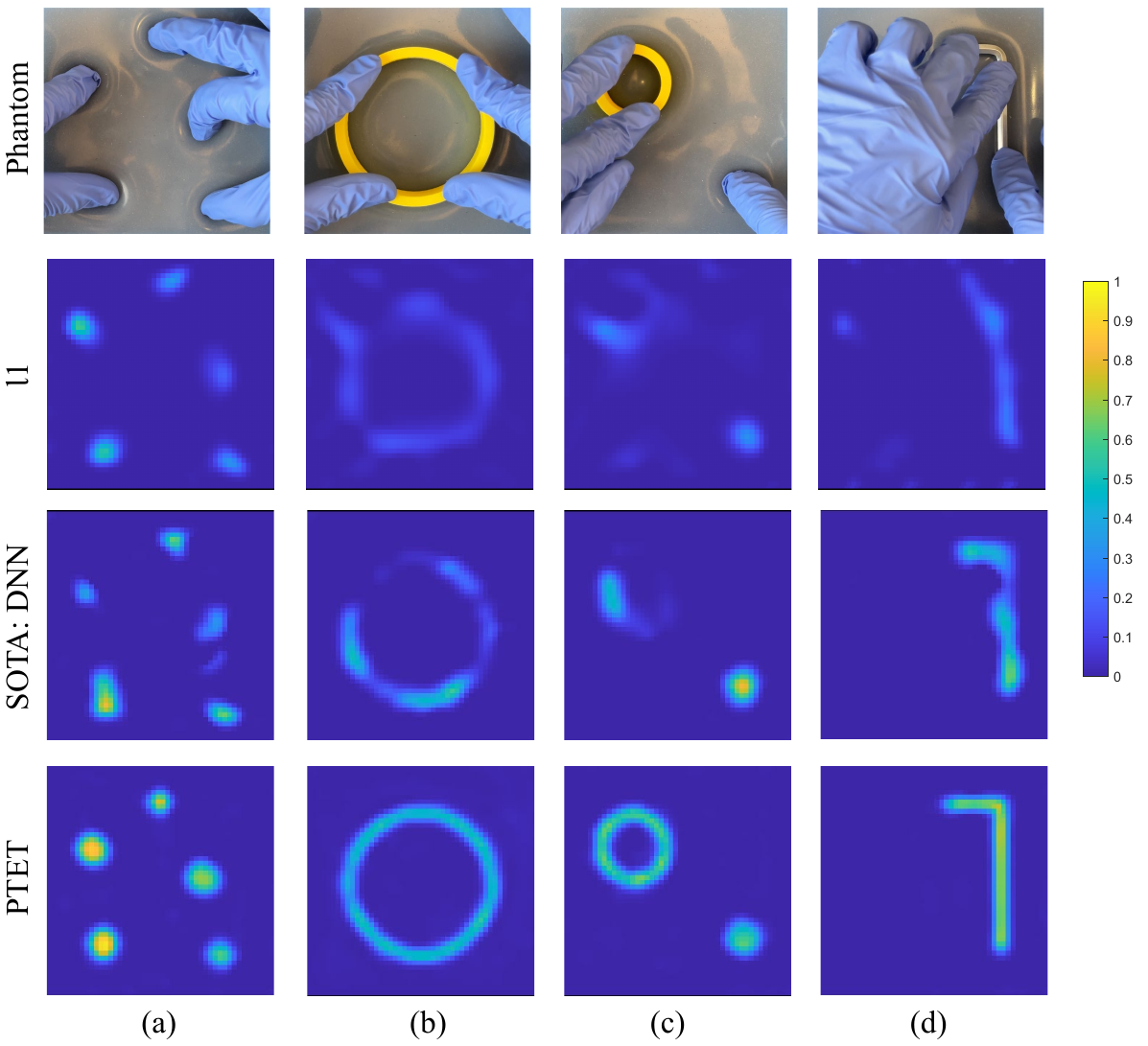}}
\caption{Tactile reconstruction for unseen real-world complex touches. (a) Multi-finger touch, (b) Annular touch, (c) Annular touch with a single finger, and (d) L-shaped touch.}
\label{fig-experimentrecon2}
\end{figure}

Fig. \ref{fig-experimentrecon2} presents a comparison among three methods: the $l1$ regularization reconstruction\cite{Tehrani2012}, SOTA DNN, and the proposed PTET framework.  The results highlight the following observations: In the multi-finger touch phantom(Fig. 10a), both $l1$ and the DNN failed to resolve clear touch boundaries,  while PTET reconstructed distinct, spatially localized regions corresponding to each finger. For the annular touch cases(Fig. \ref{fig-experimentrecon2}b and \ref{fig-experimentrecon2}c), the $l1$ and the DNN results showed partial or broken ring structures. In contrast, PTET successfully captured the complete annular shape, including the subtle structural difference introduced by the central finger in Fig. 10c. In the L-shaped touch phantom (Fig. \ref{fig-experimentrecon2}d), $l1$ reconstruction was severely blurred, and the DNN yielded only partially correct contours. However, PTET produced a sharply defined L-shape, demonstrating superior spatial resolution and generalization to previously unseen topologies.

These results indicate that the PTET framework's ability to generalize effectively to diverse real-world touch interactions even without specific labels. These results demonstrate the model's adaptability and robustness for practical applications. Supplementary Video 1 shows the real-time tactile reconstruction performance on selected phantoms.

\subsection{Comparison of Different Training Datasets}
To demonstrate the sim-to-real gap and advantages of using experimental data in tactile reconstruction, we fine-tuned PTET using three datasets: experiment dataset, noise-free simulation dataset and noisy simulation dataset. To ensure fairness, we only consider the circular touch. Each training dataset contains 2500 samples and the validation dataset contains 1000 samples. In addition, we used 504 annotated real-world samples to evaluate performance.

The quantitative metrics (Table \ref{tab-QuantitativeMetrics_experiment}) show that models trained on experimental data significantly outperformed those trained on simulated data or simulated data with noise across all metrics. The experimental data-trained model excelled in reconstructing the number, position, depth, and edge details of touch areas, consistently delivering superior accuracy and detail recovery. While adding noise to the simulated data slightly improved metrics such as CC and PSNR compared to noise-free simulation data, the experimental data still yielded the best overall performance in terms of precision and reconstruction quality.

\begin{figure}[t]
\centerline{\includegraphics[scale=0.45]{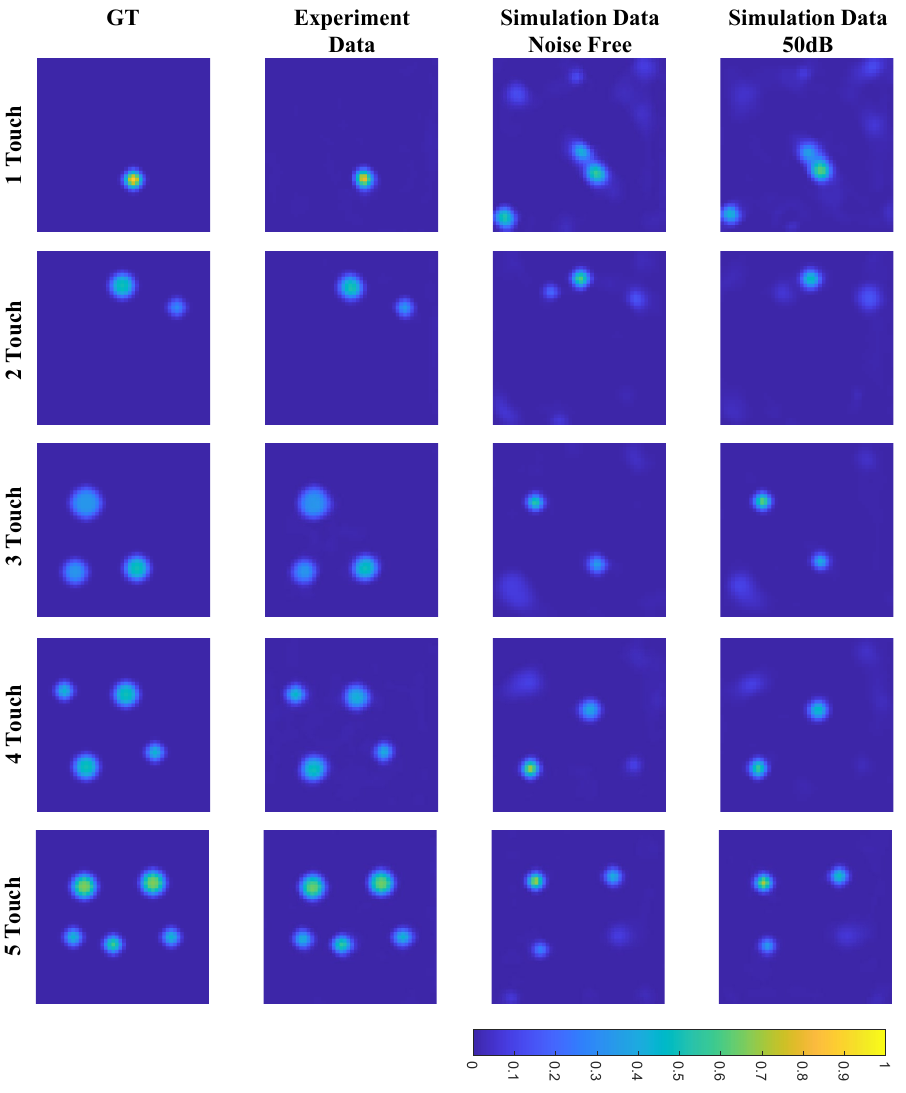}}
\caption{Reconstruction results of different Fine-tuning datasets.}
\label{fig-ComparisonDatasets}
\end{figure}

For visual comparison, Fig. \ref{fig-ComparisonDatasets} displays reconstructions of five selected phantoms from the test dataset. The results clearly illustrate the model's superior capability when trained with experimental data, accurately reconstructing the number and spatial distribution of touch points, particularly in depth and edge definition. 
\begin{table}[t]
\centering
\caption{Quantitative Metrics of Different Training Datasets}
\label{tab-QuantitativeMetrics_experiment}
\setlength{\tabcolsep}{4pt} 
\renewcommand{\arraystretch}{1.1} 
\resizebox{\columnwidth}{!}{ 
\begin{tabular}{lccc} 
\toprule
\textbf{Metrics}           & \textbf{Experiment} & \textbf{Sim (Noise Free)} & \textbf{Sim (50dB Noise)} \\ 
\midrule
RE                          & \textbf{0.3385}     & 0.8589                    & 0.8398                   \\ 
PSNR                  & \textbf{34.9600}    & 25.4645                   & 25.6551                  \\ 
CC                          & \textbf{0.8633}     & 0.4349                    & 0.4455                   \\ 
MSE ($\times 10^{-2}$)      & \textbf{0.1551}     & 0.9069                    & 0.8829                   \\ 
\bottomrule
\end{tabular}
}
\end{table}

\subsection{Human
Machine Interaction (HMI) Applications}
\begin{figure}
\centerline{\includegraphics[scale=0.4]{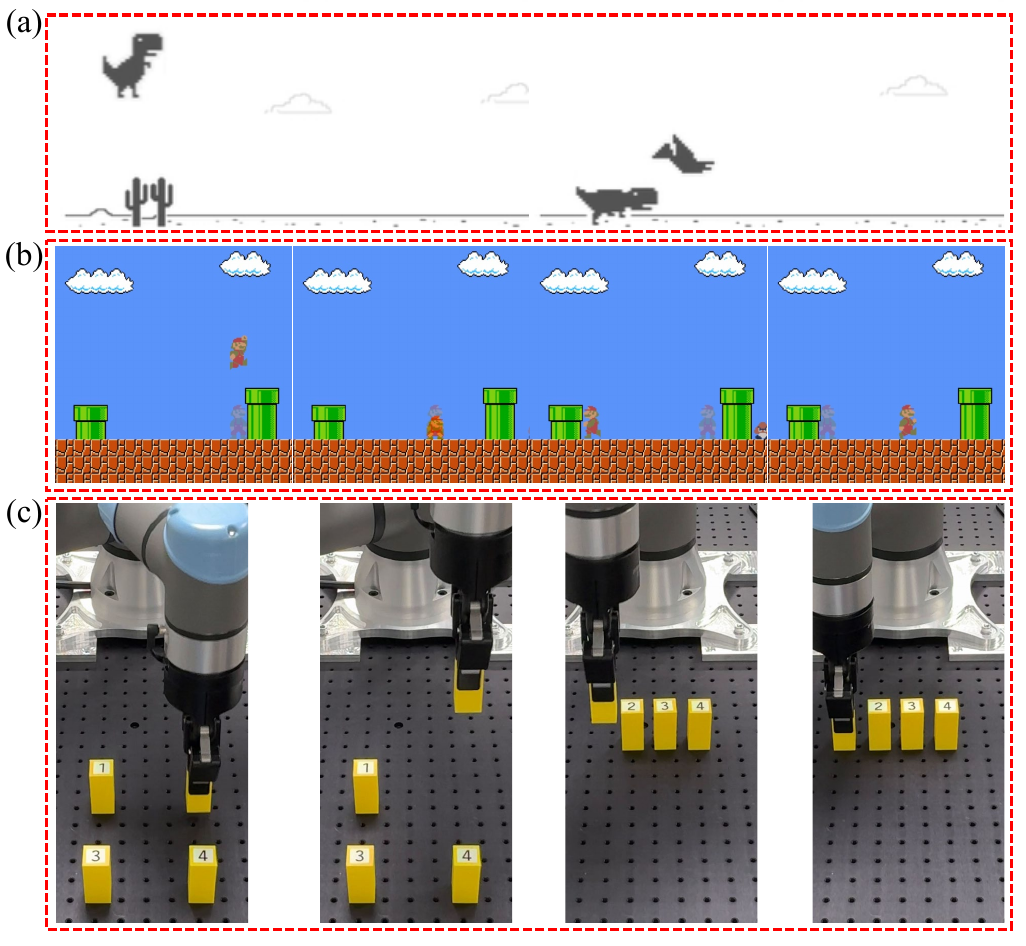}}
\caption{HMI applications (see Supplementary Video 2 for real-time control demonstrations). (a) Google Dino Game. (b) Super Mario Bros Game. (c) UR5e robotic arm.}
\label{fig-HMI}
\end{figure}

As shown in Fig. \ref{fig-HMI}, to demonstrate real-world applications, the designed tactile sensor is utilized as an HMI interface to control virtual characters in the Google Dino Game and Super Mario Bros Game. Pressing different areas triggers corresponding actions, such as jumping and crouching in Google Dino, while varying press durations enable distinct motion amplitudes, like low- and high-altitude jumps in Super Mario Bros Game. Additionally, the tactile sensor controls a UR5e robotic arm, allowing it to grab objects at specified locations and arrange them sequentially. These demonstrations highlight the sensor's ability to capture and transform subtle tactile inputs into real-time, intuitive controls for virtual environments and robotic systems. Supplementary Video 2 showcases real-time interactions with the Google Dino Game, Super Mario Bros Game, and the Universal Robots UR5e.

\section{Conclusion}\label{sec:conclusion}
This work introduces the PTET model, a novel framework for EIT-based tactile reconstruction that achieves high-resolution tactile mapping with minimal labelled real-world data through self-supervised pretraining. PTET mitigates the reliance on extensive annotated datasets and requires only 0.56\% of the annotated samples compared to the SOTA DNN model (2,500 vs. 450,000) while delivering superior tactile reconstructions. Compared to its supervised counterpart, PTET-SL, PTET achieves comparable performance using just 5\% of the annotated samples (2,500 vs. 50,000).  This is primarily due to our lightweight linear projection strategy during fine-tuning; in contrast, heavier models with more complex mapping strategies are difficult to train effectively with only 2,500 paired samples (see Supplementary Material).
Experimental results further demonstrate PTET's ability to bridge the simulation-to-reality gap and surpass the SOTA DNN model in reconstruction accuracy, achieving up to a 63.88\% improvement in overall performance.
These findings validate PTET as an effective and scalable solution for tactile perception systems, paving the way for advanced applications in HMI and robotics. In future work, we plan to explore sensor miniaturization to enable high-resolution tactile sensing in compact or embedded platforms, further expanding the application scope of PTET in wearable, prosthetic, and soft robotic systems. Beyond tactile sensing, the pre-trained encoder, based on the EIM representation, can be easily applied to other EIT sensing tasks or serve as a teacher model for developing a versatile EIT foundation model. Pretrained models with few-shot fine-tuning reduce the need for large labeled datasets in industrial measurements, enabling rapid adaptation to new conditions with minimal data. By capturing transferable features, they enhance robustness and enable cost-effective, precise, and reliable measurement with reduced downtime and calibration effort.

\bibliographystyle{IEEEtran}
\bibliography{References}
\end{document}